\documentclass[10pt,twocolumn,letterpaper]{article}

\usepackage{cvpr}
\usepackage{times}
\usepackage{epsfig}
\usepackage{graphicx}
\usepackage{amsmath}
\usepackage{amssymb}

\usepackage{booktabs}
\usepackage{algorithm2e}
\usepackage{tabularx}
\usepackage[OT1]{fontenc}

\cvprfinalcopy 

\def\cvprPaperID{****} 

\begin{document}

\title{Deformable Tube Network for Action Detection in Videos}

\author{Wei~Li, Zehuan~Yuan, Dashan~Guo, Lei~Huang, Xiangzhong~Fang and~Changhu~Wang}

\maketitle

\begin{abstract}
We address the problem of spatio-temporal action detection in videos. Existing methods commonly either ignore temporal context in action recognition and localization, or lack the modelling of flexible shapes of action tubes. In this paper, we propose a two-stage action detector called Deformable Tube Network (DTN), which is composed of a Deformation Tube Proposal Network (DTPN) and a Deformable Tube Recognition Network (DTRN) similar to the Faster R-CNN architecture. In DTPN, a fast proposal linking algorithm (FTL) is introduced to connect region proposals across frames to generate multiple deformable action tube proposals. To perform action detection, we design a 3D convolution network with skip connections for tube classification and regression. Modelling action proposals as deformable tubes explicitly considers the shape of action tubes compared to 3D cuboids. Moreover, 3D convolution based recognition network can learn temporal dynamics sufficiently for action detection. Our experimental results show that we significantly outperform the methods with 3D cuboids and obtain the state-of-the-art results on both UCF-Sports and AVA datasets.
\end{abstract}

\section{Introduction}
Action detection aims to localize actions in a long video in both space and time. In another word, we are given a video and should not only recognize actions but also find out their corresponding starting time, ending time and spatial bounding boxes of subjects. As a key ingredient for video understanding, we can easily apply it in several practical scenarios such as the surveillance and content-based retrieval. Therefore, this problem has been widely explored in recent years ~
\cite{lapeviccv2007,cao2010cvpr,Rohrbach2016,Kalogeiton_2017_ICCV,Hou_2017_ICCV,RTPReccv2018}. In particular, as deep learning techniques demonstrated a remarkable success in object detection ~\cite{renNIPS15fasterrcnn,Redmon_2016_CVPR,weiliu16ssd}, researchers followed similar pipelines (e.g., Faster R-CNN ~\cite{renNIPS15fasterrcnn}) and obtained much progress in terms of both localization and recognition. However, accurate action detection remains an open problem because it poses new challenges compared with object detection and action recognition. Crucially, bounding boxes in action detection are required to be associated with their corresponding actions, which should be inferred along time. In addition, the locations of the same actions should be consistent across adjacent frames to guarantee that humans undergo a complete action. 
 

These challenges require that actions should be represented in fine-grained granularity instead of the entire video so that they can be accurately localized.
Several previous methods ~\cite{Gkioxari_2015_CVPR,peng2016multi} processed each frame independently and recognized actions over regions  generated by selective search \cite{Uijlings2013}, region proposal network ~\cite{renNIPS15fasterrcnn} and so on. Then per-frame detections are linked to keep temporal consistency. Obviously, these methods failed to model temporal dynamics of actions which will help localization and recognition certainly. Recently, in order to include temporal information, researchers extended 2D region proposal network to generate 3D action cuboid proposals, which are then exploited to recognize actions and further refine their locations in each frame. However, the regular 3D cuboids are not able to model the flexibility of actions where the spatial positions and scales of humans might vary significantly along time though anchor cuboids are introduced to represent volumes of different scales and aspect ratios as opposed to anchor boxes in object detection. In a word, action proposals are deformable along time.


In order to overcome these shortcomings, we propose a Deformable Tube Network (DTN) for accurate action detection in videos. The network is composed of a Deformable Tube proposal Network (DTPN) and a Deformable Tube Recognition Network (DTRN), which are arranged in the same way as the standard object detection framework Faster R-CNN, as opposed to RPN and recognition network, respectively. In DTPN, action proposals are generated online by linking per-frame region proposals to form deformable region tubes. Based on these deformable tube proposals, a novel recognition network DTRN are followed to perform action recognition and location regression. DTRN consists of several 3D convolution layers to extract spatio-temporal representations. Simultaneously, skip connections are interleaved to maintain the original spatial information to enhance location boundaries. Our proposed network shows two main advantages: 1) deformable action proposals offer enough flexibility to model their variations along time, and 2) the recognition network based on 3D convolution enable us to include temporal context to improve action detection.   


In summary, our main contributions are threefold:
\begin{itemize}
\item We introduce a novel deformable tube network to perform action detection based on deformable action proposals. 
\item We propose a fast linking algorithm so that action proposals can be generated fastly.
\item We carefully design a deformable proposal recognition network to recognize actions and regress their positions.
\end{itemize}
Our experiments show that the proposed DTN model significantly outperforms other methods based on 3D cuboid proposals. Simultaneously, considering one single model and feature, DTN achieves the state-of-the-art results on UCF Sports ~\cite{ucf_sports_1,ucf_sports_2}  and  AVA dataset~\cite{Gu_2018_CVPR} . 

The reset of the paper is organized as follows. In section~\ref{sec:related_work}, we review some related works on object detection, action recognition and action detection. In section~\ref{sec:overview}, we introduce of our network structure, including DTPN and DTRN in detail. In section~\ref{sec:experiment}, we show our experiment results on UCF-Sports and AVA datasets to validate the effectiveness of our proposed model. In section~\ref{sec:discussion}, we give a further analysis of our fast tube linking algorithm and we conclude our paper in section{conclusion}
\section{Related Work}
\label{sec:related_work}

We will introduce several previous works on action detection in this section. In addition, action detection is also much related to object detection and action recognition. A lot of researches are inspired by the methodologies of object detection~\cite{DPM_ros,renNIPS15fasterrcnn,Lin_2017_ICCV,tmm4,tmm5} and the advance in action recognition~\cite{tmm1,Simonyan14b,tmm2,tmm3,Donahue_2015_CVPR,Tran_2018_CVPR}. Therefore, we will walk through these three directions. 

\subsection{Object Detection}
Recently, object detection also benefits a lot from deep learning techniques. Girshick \etal firstly proposed Region CNN (R-CNN) \cite{Girshick_2014_CVPR} in  object detection, which achieves significant improvement compared with the traditional methods ~\cite{DPM_ros,Wang_2013_ICCV,Fidler_2013_ICCV_Workshops} by introducing a convolutional neural network. Then Ren \etal\cite{renNIPS15fasterrcnn} overcame the bottleneck of speed to generate object proposals by introducing a Region Proposal Network (RPN). RPN utilizes densely sampled anchor boxes to generate top-scored object proposals. Object features extracted by ROI Pooling are then input to a recognition network for future classification and location regression. In order to accelerate this process, one-stage detectors such as SSD \cite{weiliu16ssd} and YOLO \cite{Redmon_2016_CVPR}, directly classify and regress anchor boxes in one pass without RPN to propose potential bounding boxes. In order to model the scale variations of objects, SSD assigns anchor boxes to the feature maps of different levels. Similarly, FPN \cite{Lin_2017_CVPR} builds feature pyramids by a top-down architecture with lateral connections to integrate multi-scale representations. Li \etal \cite{Li_2018_ECCV} further integrated the feature pyramid structure with a specially designed backbone for object detection. 
To address the imbalance between foreground and background instances, Lin \etal \cite{Lin_2017_ICCV} designed a novel \textit{focal loss} to automatically reduce the contribution of easy examples and focus on hard examples. Later, Mask R-CNN \cite{He_2017_ICCV} extends Faster R-CNN with a mask prediction branch to train object detection and instance segmentation in parallel.

In this paper, we follow the two-stage detection framework. However, we adapt it to action detection by extending recognition and region proposal network.

\subsection{Action Recognition}

Early approaches for action recognition mainly rely on hand-crafted features such as HOF \cite{hof} and IDT \cite{Wang2013idt}. However, they are still unable to represent abundant information in videos in spite of complex designs. 
To increase the capacity of features, several convolutional neural networks are introduced to learn video representations in an end-to-end way.  
Typically, Simonyan \etal introduced a two-stream architecture \cite{Simonyan14b} with RGB frames and optical flows as inputs to each stream respectively. The architecture process motion information and spatial appearance in parallel and fuses the classification scores of both streams to obtain a final prediction. However, extracting optical flows is time-consuming and the network still has limited capacity to capture temporal information. Trans \etal proposed a 3D convolutional neural network, where several 3D convolution operators are interleaved to learn spatio-temporal features directly. Although 3D convolution avoids the explicit extraction of optical flows, it would introduces more parameters and computation. Therefore, a lot of variants are proposed to tackle these challenges such as Pseudo 3D~\cite{Qiu_2017_ICCV_Pseudo3d} and R2+1D~\cite{Tran_2018_CVPR}. On the other hand, Donahue \etal leveraged LSTM to integrate CNN features along time~\cite{Donahue_2015_CVPR}. 
In our work, we adopt a carefully designed convolutional neural network with skip connections for action detection.

\subsection{Action Detection}

Compared to action recognition, action detection requires accurate boundaries regression. A natural way is to follow the standard sliding window strategy in object detection ~\cite{lapeviccv2007,cao2010cvpr,Rohrbach2016}. The main difference mainly lies in the feature selection and the method to generate candidate action proposals. 
For example, Rohrbach \etal \cite{Rohrbach2016} generated multiple candidate segments by sliding windows and perform recognition over dense trajectories and human pose features. 
Lan \etal~\cite{laniccv2011} made use of figure-centric visual word features to represent actions. 

As the performance of object detection went up, most recent approaches turned to link frame-level object detections to form action tubes. Based on visual and motion cues from two-stream network, Gkioxari and Malik \cite{Gkioxari_2015_CVPR} classified region proposals generated by selective search, which are then linked to action tubes along time for temporal consistency. Weinzaepfei \etal \cite{Weinzaepfel_2015_ICCV} proposed to track high-scoring proposals using a tracking-by-detection approach. Saha \etal \cite{Saha2016DeepLF} fused appearance and motion detection boxes based on estimated action scores and their spatial overlaps between each other, and constructed spatio-temporal action tubes with a two-pass dynamic programming method. Peng and Schmid \cite{peng2016multi} replaced selective search with region proposal network and embedded a multi-region scheme into their two-stream classification network. Singh \etal \cite{singh2016online} introduced a real-time action localization method with a SSD object detector and an online linking algorithm. All these methods rely much on frame-level human detections. However, distinguishing actions based on single frames are difficult without considering temporal dynamics. 

To further include temporal dynamics for action recognition and localization, Kalogeiton \etal \cite{kalogeiton17iccv} came up with anchor cuboids to generate action proposals directly, which encode enough spatio-temporal information for action recognition. Hou \etal \cite{Hou_2017_ICCV} further generalized Region-of-Interest (RoI) pooling layer to 3D Tube-of-Interest (ToI) pooling layer. Although these approaches built on anchor cuboids offer an opportunity to integrate temporal dynamics for action recognition compared with those connecting frame-level human detections, action tubes are too deformable to be modelled by regular 3D volumes in practice.  
In contrast, we propose a novel detection network, which generates deformable action tube candidates by a fast linking algorithm and perform recognition and regression over these proposals. RTPR \cite{RTPReccv2018} is probably most similar to us in spirit, which also link region proposals to action tubes and perform action recognition by LSTM. However, in their work, action localization are regressed recurrently and did not consider global temporal information to enhance detections in each frame. Instead, we design a fully convolutional neural network to perform recognition and regression as a whole over generated deformable action tubes. 

\section{Overview of the approach}
\label{sec:overview}

\begin{figure*}
\begin{center}
\includegraphics[width=1.0\linewidth]{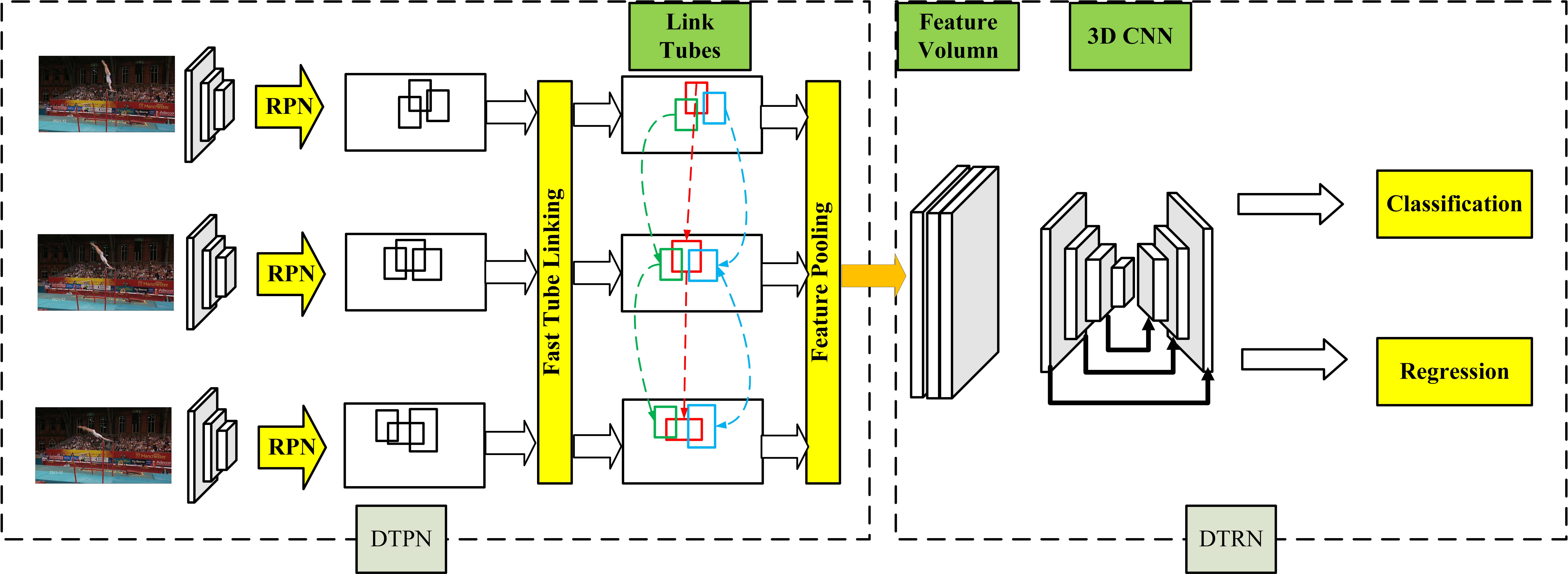}
\end{center}
   \caption{The overall architecture of our proposal two-stage action localization model with DTPN and DTRN. We link per-frame proposals into deformable candidate tubes in our DTPN, which can capture actors in a more flexible way. Our DTRN can capture spatio-temporal correlations with 3D Convolution to perform recognition and bounding box regression.}
\label{fig:architecture}
\end{figure*}

Our proposed deformable tube network is composed of a deformable tube proposal network (DTPN) and a deformable tube recognition network (DTRN). An overview of our framework is shown in Figure~\ref{fig:architecture}. 
To generate deformable actor-centric tube proposals, we decompose it into two separate processes. Firstly, we obtain region proposals of high quality for each frame using a standard RPN. Then, we leverage a fast tube linking algorithm (FTL) to link per-frame proposals into deformable candidate tubes. Compared with anchor cuboids, deformable actor-centric tubes can capture actors in a more flexible way but also maintain the spatio-temporal information. Then DTRN is carefully designed to classify and refine candidate tube proposals.

\subsection{Deformable Tube Proposal Network}
\label{sec:1}

The tube proposal network aims to generate deformable spatio-temporal action proposals. 
The whole network is composed of a standard RPN and one linking layer.  RPN  generates a set of candidate actor region proposals for each frame.  
In the following linking layer, we employ a fast proposal linking algorithm to generate deformable action tubes. 

\subsubsection{Region Proposals Generation}

In our experiment, RPN takes as input video frames and generates 1k actor proposals per frame. It is worth noting that RPN does not consider the temporal context across frames and performs region proposal generation independently for each frame. 
We follow the standard setting during training. Specifically, several anchor bounding boxes are first generated according to predefined aspect ratios and scales. Then the overlap score with any groundtruth is estimated for each and every anchor. Anchor are considered as positives if their maximum overlap scores are higher than 0.7 
regardless of action labels and the rest are assigned negative labels. Besides classification, positive anchors are also used to regress their corresponding groundtruth locations. 

\subsubsection{Deformable Tube Proposals}

For each and every frame in a video clip, RPN outputs a set of region proposals each with a confidence score to represent how likely it belongs to an actor. Inspired by ~\cite{RTPReccv2018}, we link these generated regions with large overlaps across frames to form deformable action tubes by iteratively running a dynamic programming algorithm, which are then fed to the following recognition network to perform classification and further regression. 
Instead of considering overlap ratios during the tube construction as in ~\cite{RTPReccv2018}, we adopt a hard thresholding strategy to accelerate the process. Nevertheless, generating proposals serially is still expensive in the entire detection pipeline. Hence, we improve the linking algorithm a lot to further reduce the generation computation cost.

Given a video clip with continuous $T$ frames and RPN generates $N^t$ proposals $\{r_t^1,...,r_t^{N^t}\}$ in the $t$-th frame, our goal is to generate $M$ deformable tube proposals  $\{P_1^{1\rightarrow{}T},...,P_M^{1\rightarrow{}T}\}$ with largest action scores.

For any tube $P_i^{1\rightarrow{}T}=\{r_i^1,...,r_i^T\}$, its action score represents how likely this tube is regarded as an action and is formulated as 
\begin{equation}
S(P_i^{1\rightarrow{}T})=(\sum_{t=1}^T a_i^t) + T * L(P_i^{1\rightarrow{}T})\,
\label{eq:1}
\end{equation}{}
where $a_i^t$ is the objectness score of $r_i^t$ output by RPN and $L(P_i^{1\rightarrow{}T})$ is a hard thresholding score of $P_i^{1\rightarrow{}T}$ defined as
\begin{equation}
L(P_i^{1\rightarrow{}T})=
\begin{cases}
    1,& \text{if } \forall t, iou(r_i^t, r_i^{t+1}) > \tau \\
    0, & \text{otherwise}
\end{cases}
\end{equation}
We can see that the tubes with high action scores prefer the region proposals with large overlaps across frames and high confidence scores.

As known, finding the tube proposal with the maximum score $S$ can be converted into a typical dynamic programming problem and addressed by Viterbi algorithm~\cite{viterbi}.  In order to search for $M$ largest action proposals, we adopt a greedy strategy. Firstly, the tube with the largest score is generated.  Then we remove all region proposals in it and find next path with the maximum action score in the remaining proposals. The whole process terminates when all $M$ tube proposals are found or no legal region candidates can be linked. Assuming that each frame has $N$ candidate regions averagely, the time complexity of the whole process is $M\times T\times N^2$. In our practical experiments, we found that this greedy generation took too long as the Viterbi algorithm in each step cost a lot. Therefore, we improve the dynamic programming algorithm to boost the deformable tube generation. 

Our main modifications lie in two algorithm details.

\textbf{Hard Thresholding} One hand, since there a large amount of region proposals in each frame and actors usually move smoothly, we only link proposals to those with large overlaps in follow frames, which reduces much searching space and thus increase the efficiency. Specially, given an uncompleted path $P_j^{1\rightarrow{}t-1}=\{r_j^1,...,r_j^{t-1}\}$ and $N^t$ proposals $R^t=\{r_t^1,...,r_t^{N^t}\}$ in  next frame, we only choose the region proposals $r_t  \in R^t$ with IoU scores with $r_j^{t-1}$ higher than $\tau$ to extend the tube. As a result, the time complexity is reduced from $M \times T \times N^2$ to $ M\times T \times N\times Q$ where $Q$ is the average number of legal proposals in next frame. Most importantly, the tube proposal generated by this way is still the global optimal solution due to our hard thresholding strategy.  

\textbf{Top-K Selection} Additionally, maintaining and updating the maximum action scores of incomplete proposals ending at all region proposals in intermediate frames are unnecessary and expensive. Based on the observation that RPN performs relatively reliable in each frame and the tube with the largest confidence always scores high in every step, we only select top-$K$ uncompleted path $P_i^{1\rightarrow{}t}$ as candidates tubes to extend at any step $t$ (See Algorithm ~\ref{alg:1}). We will discuss the influence of $K$ in the experiment section. Consequently, we reduce the time complexity of the whole generation process from $M\times T\times N^2$ to $M\times T\times Q \times K$ which significantly accelerates action detection.       

\begin{algorithm}
\label{alg:1}
\SetKwInOut{Input}{Input}
\SetKwInOut{Output}{Output}
\Input{Region Proposals $R_t=\{r_t^1,...,r_t^{N^t}\}$ in each frame $t$, adjustable parameter $K$.}
\Output{Tube proposals ${\Omega}=\{P_1^{1\rightarrow{}T},...,P_M^{1\rightarrow{}T}\}$.}

\For{$m = 1$ to $M$}{
    \For{$t = 1$ to $T$}{
        \uIf{$t = 1$}{
            $\hat{P}^{1\rightarrow{}{t-1}}\gets \{\hat{P}_1^{1\rightarrow{}1},...,\hat{P}_K^{1\rightarrow{}1}\} \gets$ top-$K$ $r_i^1$ by $\alpha_i^1$
        }
        \uElse{
            \For{$r_i^t$ in frame $t$}{
                ${\overset{\ast}{P}}_i^{1\rightarrow{}t} \gets \mathrm{argmax}_{\hat{P}_k} S(\hat{P_k}^{1\rightarrow{}{t-1}} \cup r_i^t)$
            }
            $\hat{P}^{1\rightarrow{}t}\gets\{\hat{P}_1^{1\rightarrow{}t},...,\hat{P}_K^{1\rightarrow{}t}\} \gets$ top-$K$ \\ ${\overset{\ast}{P}}_i^{1\rightarrow{}t}$ by $S$ in Eq~\ref{eq:1}
        }

    }
    $R_t \gets R_t \setminus \{{\overset{\ast}{P}}_k^{1\rightarrow{}T}\}_{best}$ \\
    $\Omega \gets \Omega \cup \{{\overset{\ast}{P}}_k^{1\rightarrow{}T}\}_{best}$ \\
}
\caption{Fast tube linking algorithm. $\hat{P}^{1\rightarrow{}t}$ represent incomplete tubes ending at $t$ frame. $\{\cdot\}_{best}$ aims to  select tubes with the largest action score.}
\end{algorithm}


\subsection{Deformable Tube Proposal Recognition}

When $M$ deformable tube proposals are available, the tube proposal recognition network aims to classify them correctly and refine their locations in each frame. Similar to Fast RCNN, the network takes video frames and deformable tube proposals as input. 
The network firstly processes video frames by a backbone convolutional neural network and then gets multi-scale feature maps for each frame. Then a feature pooling layer outputs a 3D feature volume for each deformable tube proposal, which are further fed into our designed tube recognition network for action classification and location regression. 

\textbf {Tube Feature Pooling} To form the representation for a tube, we firstly exploit ROI Pooling to extract region features independently based on the previous feature maps and then these representations are concatenated along the tube. Specifically,  given any tube $P\in \mathbb{R}^{T*4}$ with $T$ frames where 4 denotes the region positions of $P$ in each frame as in Faster R-CNN~\cite{renNIPS15fasterrcnn}, ROI Pooling outputs a feature volume with a fixed spatial extent $h\times w$ (e.g. $14\times14$). By stacking features along T frames, we obtain a $T\times h\times w$ representation. 

\textbf {Tube Recognition Network} Our recognition network takes tube representations as input and performs recognition and bounding boxes regression. 
The entire architecture follows the design of U-net~\cite{unet} with extra skip connections illustrated in Table~\ref{tab:1}, where inputs undergo a series of spatial and temporal convolutions to enhance the interaction along tubes and model motion dynamics.  In particular for temporal convolution layers, we use the kernel $3 \times 1 \times 1$ without padding to reduce the temporal dimension. 
The same amount of deconvolution layers are followed to recover the original spatial and temporal resolutions, which are used for bounding box regression and classification later. It is worth noting that we also adopt skip connections to connect the previous convolution layers with their symmetric deconvolution layers, which aims to help regression by adding low-level features. 

To get final predictions, two sibling layers are followed to estimate action classes and positions respectively. The first outputs a probability distribution for each tube, $p=(p_0,...,p_C)$, over $C+1$ categories. For the $t$-th region $b_t=(x_t,y_t,w_t,h_t)$ of the tube, the second sibling layer outputs bounding box regression offsets $\Delta_t^c=(\Delta x_t^c,\Delta y_t^c,\Delta w_t^c,\Delta h_t^c)$ for each of the $C$ action classes. The corresponding regression target $\widetilde{\Delta}_t^c=(\widetilde{\Delta} x_t^c,\widetilde{\Delta} y_t^c,\widetilde{\Delta} w_t^c,\widetilde{\Delta} h_t^c)$ is designed as
\begin{equation}
    \widetilde{\Delta} x_t^c=(x_t-x_t^*)/x_t^*
\end{equation}
\begin{equation}
    \widetilde{\Delta} y_t^c=(y_t-y_t^*)/y_t^*
\end{equation}
\begin{equation}
    \widetilde{\Delta} w_t^c=log(w_t/w_t^*)
\end{equation}
\begin{equation}
    \widetilde{\Delta} h_t^c=log(h_t/h_t^*)
\end{equation}
where $b_t^*=(x_t^*,y_t^*,w_t^*,h_t^*)$ is the assigned groundtruth of $b_t$ with class $c$.

Given these definitions, our multi-task loss for each tube is defined as:
\begin{equation}
\begin{aligned}
    \mathcal{L}(p,b_t,b_t^*,c)=&\mathcal{L}_{cls}(p,c)\\
        &+\mathbf{1}(c>0)\sum_{t=1}^{T}L_{reg}(\Delta_t^c,\widetilde{\Delta}_t^c)
\end{aligned}
\end{equation}
where $\mathbf{1}$ is an indicator function, ${L}_{cls}(p,c)$ represents cross entropy loss  with true class $c$ and $L_{reg}(\Delta_t^k,\widetilde{\Delta}_t^k)$ is a regression loss formulated as
\begin{equation}
    L_{reg}(\Delta_t^c,\widetilde{\Delta}_t^c)=\sum_{i\in \{x,y,w,h\}}SmoothL1(\Delta i_t^c, \widetilde{\Delta} i_t^c)
\end{equation}


\begin{table}
\caption{Detailed network architecture of DTRN.}
\begin{center}
\begin{tabular}{|p{2.0cm}p{1.2cm}p{1.2cm}p{2.4cm}|}
\hline
name & kernel & stride & output\\
\hline\hline
Input & - & - & $5\times256\times14\times14$ \\
Conv1 & $1\times3\times3$ & $1\times2\times2$ & $5\times512\times7\times7$ \\ 
Conv2 & $3\times1\times1$ & $1\times1\times1$ & $3\times512\times7\times7$ \\
Conv3 & $3\times1\times1$ & $1\times1\times1$ & $1\times512\times7\times7$ \\
Deconv3 & $3\times1\times1$ & $1\times1\times1$ & $3\times512\times7\times7$ \\
Skip3 & - & - & $3\times512\times7\times7$ \\
Deconv2 & $3\times1\times1$ & $1\times1\times1$ & $5\times512\times7\times7$ \\
Skip2 & - & - & $5\times512\times7\times7$ \\
Deconv1 & $1\times3\times3$ & $1\times2\times2$ & $5\times256\times14\times14$ \\
Skip1 & - & - & $5\times256\times14\times14$ \\
Flatten-Reg & - & - & $5\times50176$ \\
Flatten-Cls & - & - & $250880$\\
\hline
\end{tabular}
\end{center}
\label{tab:1}
\end{table}


\section{Experiments}
\label{sec:experiment}
\subsection{Datasets}
Unlike image dataset for object detection, large scale video datasets for spatial-temporal localization is much harder to collect. We evaluate our proposed model in two action localization datasets: UCF-Sports \cite{ucf_sports_1,ucf_sports_2} and AVA \cite{Gu_2018_CVPR}. UCF-Sports dataset consists of 150 short videos with 10 different actions. The bounding box annotations are available for all frames. We follow the dataset split of \cite{iccv2011lan} with 103 videos for training and 47 videos for testing.
AVA is a newly published challenging dataset. It densely annotates 80 atomic visual actions in 430 video clips with 15 minutes and nearly 900 key frames are annotated based on 3-seconds video segments centered on these key frames. Each bounding box of key frames has multiple action labels. We follow the experiment setup of \cite{Gu_2018_CVPR}. It only uses classes that have at least 25 instances, which results in a total of 210,634 training and 57,371 validation examples on 60 classes. 

\subsection{Experiment Setup}
\label{sec:setup}
\subsubsection{Implementation Details}
\label{sec:detail}
We implement our method based on the MXNet toolbox \cite{Chen2015MXNetAF}. When training RPN, we assign positive labels to anchor boxes with an IoU higher than 0.7 or has the highest IoU with any groundtruth.  All remaining anchors are considered as negatives. We keep top $N=1000$ RPN proposals after NMS operation as candidate region proposals for tube linking. Our FTL links these region proposals to generate $M=200$ tube proposals with a IoU threshold $\tau =0.3$. One tube proposal is assigned to a positive label if the average IoU  of boxes over all frames with ground truths is higher than 0.5 and the rest are associated with negative labels. 
 In terms of DTRN's training, we sample 40 tube proposals per video as inputs for backpropagation, where positive tubes and negative ones keep a ratio of $1:3$. In testing stage, non-maximum suppression (NMS) with a threshold of 0.7 is applied in each frame to get the final action detection results. We add random flip to the whole sequence of frames to prevent overfitting in training. For the training strategy, we employ SGD optimizer of \cite{lecun_bp} a momentum of 0.9 and a clipnorm of 5. The learning rate is initialized with 0.0004. We experiment with 2D backbone for UCF-Sports dataset and both 2D and 3D backbone for AVA dataset. For UCF-Sports dataset, we sample 5 frames with a sampling stride of S frames. As for AVA dataset, we densely sample 5 frames in 1-second video segment centered at this key frame for 2D ResNet-101 backbone. As for 3D ResNet-50 backbone \cite{slowfast}, We choose to generate our deformable tube with region proposals of consecutive key frames. For each key frame, compared with single frame input for 2D backbone, the input for our 3D ResNet-50 backbone is 32 frames sampled from a 64-frame raw clip with a temporal stride of 2. We sample 5 video clips each centered with key frames with a sampling stride of $S$ seconds. For our DTN with 3D backbone, we follow a two-steps training with a \textbf{baseline model} which only utilizes the center video clip to predict action categories and regress box offsets of corresponding key frame as \cite{Gu_2018_CVPR}. Then we extract the feature volume of $res_4$ and finetune $res_5$ with loaded weights from baseline model due to GPU memory limit. Finally, we use mean pooling to reduce the temporal dimension of the output feature volume of each clip to 1 and stack each feature map after ROI- Pooling together as the input for our DTRN.


\begin{figure*}
\begin{center}
\includegraphics[width=1.0\linewidth]{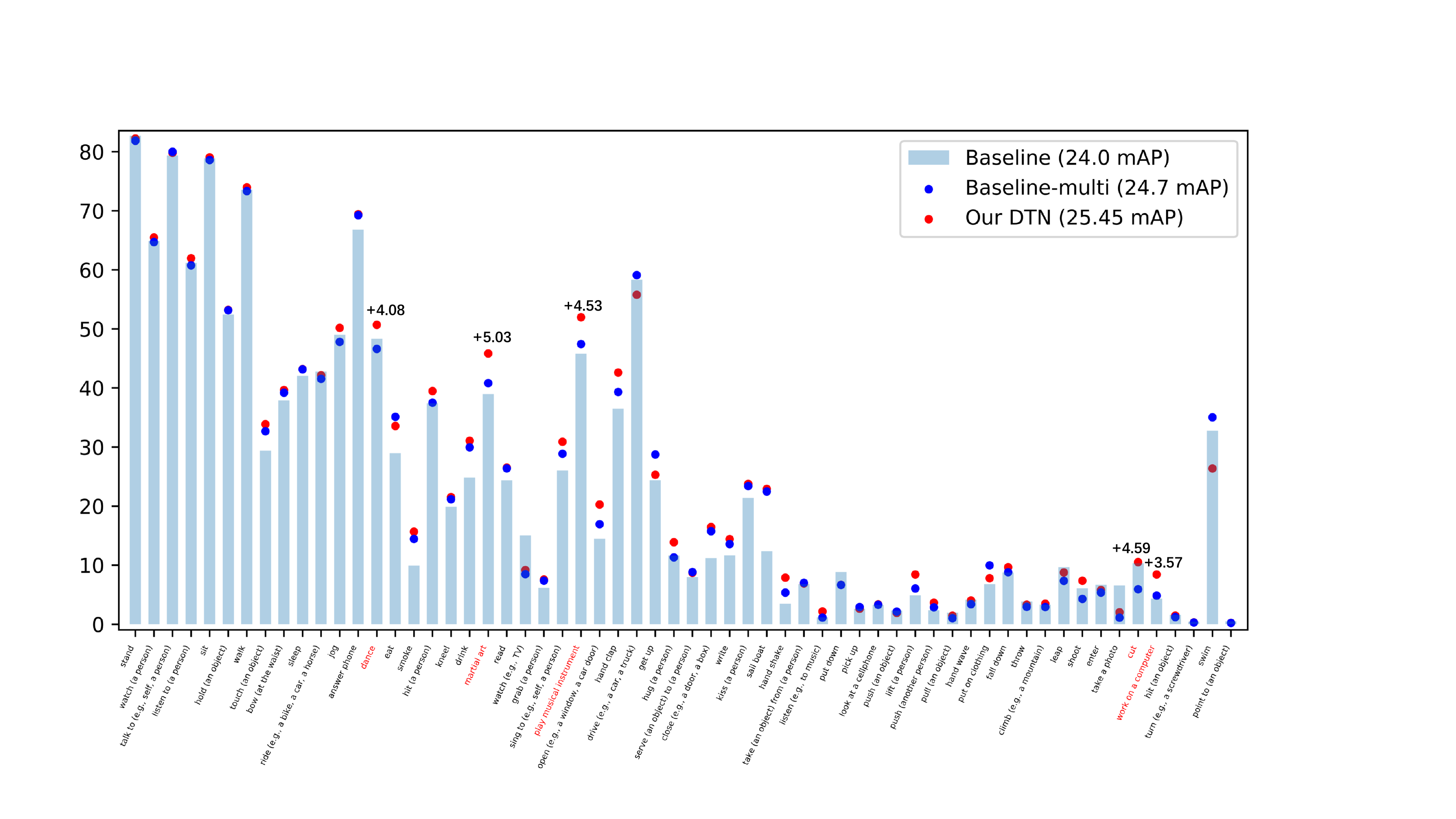}
\end{center}
   \caption{Per-category AP on AVA dataset: baseline model, baseline-multi model and Our DTN. Categories are sorted by the number of training examples. Our DTN achieves 0.75 points performance improvement compared with baseline-multi model. The highlighted categories are the 5 largest absolute gains (\textbf{black}).}
\label{fig:ava_comparison}
\end{figure*}

\subsubsection{Evaluation Metrics}
To evaluate our model performance, we adopt \textit{frame-mAP} and \textit{video-mAP} as our evaluation metrics. 
An action tube or a frame is considered as positive if the IoU with the groundtruth annotation is greater than a threshold and the action prediction is correct. Specially, we utilize both \textit{frame-mAP} and \textit{video-mAP} as evaluation metrics for UCF-Sports. Since AVA dataset is annotated with multiple labels for each bounding box, we only use \textit{frame-mAP} for AVA dataset.

\begin{table}

\caption{Runtime (s) comparison of different linking algorithm.}
\begin{center}
\begin{tabular}{lccccc}
\hline
& N=300 & N=500 & N=700 & N=1000 \\
\hline
Viterbi & 8.27 & 34.04 & 70.4 & 122.7 \\
Viterbi + HT & 2.63 & 5.3 & 7.72 & 12.0 \\
Viterbi + HT + TS & \textbf{0.461} & \textbf{0.560} & \textbf{0.641} & \textbf{0.669} \\
\hline
\end{tabular}
\end{center}
\label{tab:11}
\end{table}

\begin{table}
\caption{Comparisons of variants with different tube linking algorithm on UCF-Sports dataset.}
\begin{center}
\begin{tabular}{lcc}
\hline
  & \textit{frame-mAP} & \textit{video-mAP} \\
  & $(\sigma=0.5)$ & $(\sigma=0.2)$ \\
\hline
Viterbi & 91.9 & \textbf{99.0} \\
Viterbi + HT & 92.3 & 98.3 \\
Viterbi + HT + TS & \textbf{93.08} & 98.8 \\
\hline
\end{tabular}
\end{center}
\label{tab:12}
\end{table}

\subsection{Ablation Study}
\subsubsection{Runtime Analysis}
As mentioned in Section \ref{sec:1}, our modifications of linking algorithm lie in two details, hard thresholding (HT) and top-K selection (TS). We reduce the time complexity of the linking algorithm from $M\times T\times N^2$ to $M\times T\times Q\times K$. Table~\ref{tab:11} shows the actual runtime of different linking algorithm with N region proposals. The frame count of each linked tube if fixed to 5 for all linking algorithm.

We can achieve over 180x acceleration with our two modifications (HT and TS) when $N=1000$ and nearly 18x acceleration when $N=300$.

To further evaluate the quality of our linked tubes, we compare the performance with different linking algorithm on UCF-Sports dataset in Table~\ref{tab:12}. We can achieve slightly worse or even better result with our two modifications, which verifies the effectiveness of our fast linking algorithm. We will further quantify the effect of TS in Section~\ref{sec:ts}.

\subsubsection{Linked Tubes \& Anchor Cuboids} As discussed in Section~\ref{sec:1}, our TDN can generate deformable actor-centric tube proposals. Compared with anchor cuboids, our deformable candidate tubes can capture actors in a more flexible way. We will give a qualitative visualization to illustrate the advantage of our linked tubes in Section~\ref{sec:2}. In this section, we quantitively evaluate the performance with linked tubes or anchor cuboids on both UCF-Sports and AVA datasets.

For UCF-Sports dataset, we choose top-200 region proposals of the center frame and replicate it 5 times to form anchor cuboids. As shown in Table~\ref{tab:13}, our DTN with linked tubes outperforms its counterpart with anchor cuboids in most categories, especially in \textit{Swing2 (Swing-SideAngle)}, \textit{Walk}, \textit{Run} and \textit{Riding}. These are the categories with relatively fast motion of actors.

\begin{table*}
\caption{Comparison results of variant model with anchor cuboids or linked tubes on UCF-Sports dataset.}
\centering
\begin{tabular}{p{2cm}p{0.6cm}p{0.6cm}p{0.8cm}p{0.9cm}p{0.9cm}p{0.5cm}p{0.8cm}p{0.8cm}p{0.8cm}p{0.7cm}p{1.5cm}p{1.5cm}}
\hline
  & Diving & Golf & Kicking & Lifting & Riding & Run & SkateB. & Swing1 & Swing2 & Walk & frame-mAP & video-mAP\\
\hline
 Anchor Cuboids & 100.00 & \textbf{99.62} & 88.56 & 100.00 & 92.14 & 81.31 & \textbf{98.13} & \textbf{89.40} & 81.55 & 72.52 & 90.32 & 97.8 \\
 Linked Tubes & \textbf{100.00} & 95.20 & \textbf{88.62} & \textbf{100.00} & \textbf{98.53} & \textbf{90.03} & 96.98 & 77.44 & \textbf{99.36} & \textbf{84.61} & \textbf{93.08} & \textbf{98.8} \\
\hline
\end{tabular}
\centering
\label{tab:13}
\end{table*}

Compared with UCF-Sports dataset, the movement of actors in AVA dataset is relatively slow. We do all the comparisons with 3D backbone with relatively larger sampling intervals. As mentioned in Section~\ref{sec:setup}, we first train a baseline model which only utilizes the center video clip and then finetune $res_5$ stage to integrate information over the span of multiple seconds with deformable linked tubes. In order to evaluate the advantage of our deformable tubes compared with anchor cuboids, we replace the linked tubes with top-200 region proposals of the center key frame as baseline-multi model. The comparison results are shown in Table~\ref{tab:6}. We can see that our DTN obtains consistent improvements compared with baseline and baseline-multi models. Our DTN outperforms baseline model by 1.84 points when $S=2$ and baseline-multi model by 0.75 points when $S=1$, which verifies the effectiveness of our linked tubes compared with anchor cuboids on AVA dataset.

\begin{table}
\caption{Comparisons with baseline models on AVA dataset with \textit{frame-mAP}. The Iou threshold $\sigma$ is fixed to 0.5.}
\label{tab:6}
\begin{center}
\begin{tabular}{lccc}
\hline
  & $S=1$ & $S=2$ \\
  \hline
 baseline & 24.0 & 24.0  \\
 baseline-multi & 24.7 & 25.2 \\
 our DTN & \textbf{25.45} & \textbf{25.8} \\
\hline
\end{tabular}
\end{center}
\end{table}

As for the per-category AP shown in Figure～\ref{fig:ava_comparison}, our DTN outperforms baseline model in \textbf{44} of \textbf{60} categories and baseline-multi model in \textbf{50} of \textbf{60} categories. We can see the largest absolute gains for categories like ``martial art(+5.03)'', ``cut(+4.59)'', ``play musical instrument'' and ``dance(+4.08)'' when compared with baseline-multi model. Our deformable tube play a key role for such categories with large spatial movements.


\subsection{Variants of Proposed Model}
\subsubsection{Sampling Interval} For UCF-Sports dataset, we sample 5 consecutive frames with a sampling interval of $S$ frames as the input of our DTRN. The sampling interval plays a key role in our DTN, which controls the diversity of each input frame. In order to investigate the effect of sampling interval on the performance of our DTN, we increase the sampling interval from 1 to 4 to obtain 4 variants of our model. The comparison results with different sampling intervals are shown in Table~\ref{tab:8}. With larger sampling interval, our DTN can integrate more intact video contents which helps to distinguish similar action categories. We achieve consistent improvement by changing the sampling interval $S$ from 1 to 3. However, there is a performance drop by changing $S$ from 3 to 4 which can be explained by that large sampling interval can leads to inaccurate linking of region proposals when with fast motions. Based on these analyses, we fix the sampling interval to 3 for all other experiments on UCF-Sports dataset.

\begin{table}
\caption{Comparisons of variants with different sampling interval on UCF-Sports dataset.}
\begin{center}
\begin{tabular}{lcc}
\hline
  & \textit{frame-mAP} & \textit{video-mAP} \\
  & $(\sigma=0.5)$ & $(\sigma=0.2)$ \\
\hline
$S=1$ & 91.2 & 97.3 \\
$S=2$ & 92.64 & 98.0 \\
$S=3$ & \textbf{93.08} & \textbf{98.8} \\
$S=4$ & 92.2 & 98.3 \\
\hline
\end{tabular}
\end{center}
\label{tab:8}
\end{table}

\subsubsection{Combination with LFB} LFB \cite{lfb2019} can be used to augment 3D Convolution Networks with supportive information extracted over the whole span of a video. It extends vanilla 3D Convolution Networks with a external long-term feature bank and a feature bank operator (e.g. Non-local operator) that computes interactions between short-term and long-term features. We argue that our deformable tube which integrates middle-term video contents is complementary to LFB. For each key frame, we choose top-2 linked tubes with highest action scores (maximum class score after sigmoid activation of all 60 classes). The feature volume before classification stage is used as the representation for this linked tube (after vectorization by flatten or global average pooling). We replace the precalculated long-term feature bank in LFB with our extracted representations. The comparison results of our DTN with or without LFB are shown in Table~\ref{tab:14}. We can see a consistent improvement when combined with LFB.

\begin{table}
\caption{Comparison results of our DTN with or without LFB on AVA dataset with \textit{frame-mAP}. The Iou threshold $\sigma$ is fixed to 0.5.}
\label{tab:14}
\begin{center}
\begin{tabular}{lccc}
\hline
  &  \textit{frame-mAP} \\
  \hline
 our DTN &  25.8 \\
 our DTN + LFB (average) & 27.2 \\
 our DTN + LFB (flatten) & \textbf{27.7} \\
\hline
\end{tabular}
\end{center}
\end{table}

\begin{table}
\caption{Comparison results with the state of the art methods on UCF-Sports dataset with \textit{frame-mAP} and \textit{video-mAP}. The Iou threshold $\sigma$ is set to 0.5 and 0.2 respectively.}
\begin{center}
\begin{tabular}{lcc}
\hline
  & \textit{frame-mAP} & \textit{video-mAP} \\
  & $(\sigma=0.5)$ & $(\sigma=0.2)$ \\
\hline
Gkioxari \etal \cite{Gkioxari_2015_CVPR} & 68.1 & \\
Weinzaepfel \etal \cite{Weinzaepfel_2015_ICCV} & 71.9 & - \\
Peng \etal \cite{peng2016multi} & 84.51 & 94.8 \\
Kalogeiton \etal \cite{Kalogeiton_2017_ICCV} & 87.7 & 92.7 \\
Hou \etal \cite{Hou_2017_ICCV} & 86.7 & 95.2 \\
He \etal \cite{He2018GenericTP} & - & 96.0 \\
Li \etal \cite{RTPReccv2018} & - & 98.6 \\
Duarte \etal \cite{Duartearxiv} & 83.9 \\
\hline
Our DTN & \textbf{93.08} & \textbf{98.8} \\
\hline
\end{tabular}
\end{center}
\label{tab:3}
\end{table}

\begin{figure*}
\begin{center}
\includegraphics[width=1.0\linewidth]{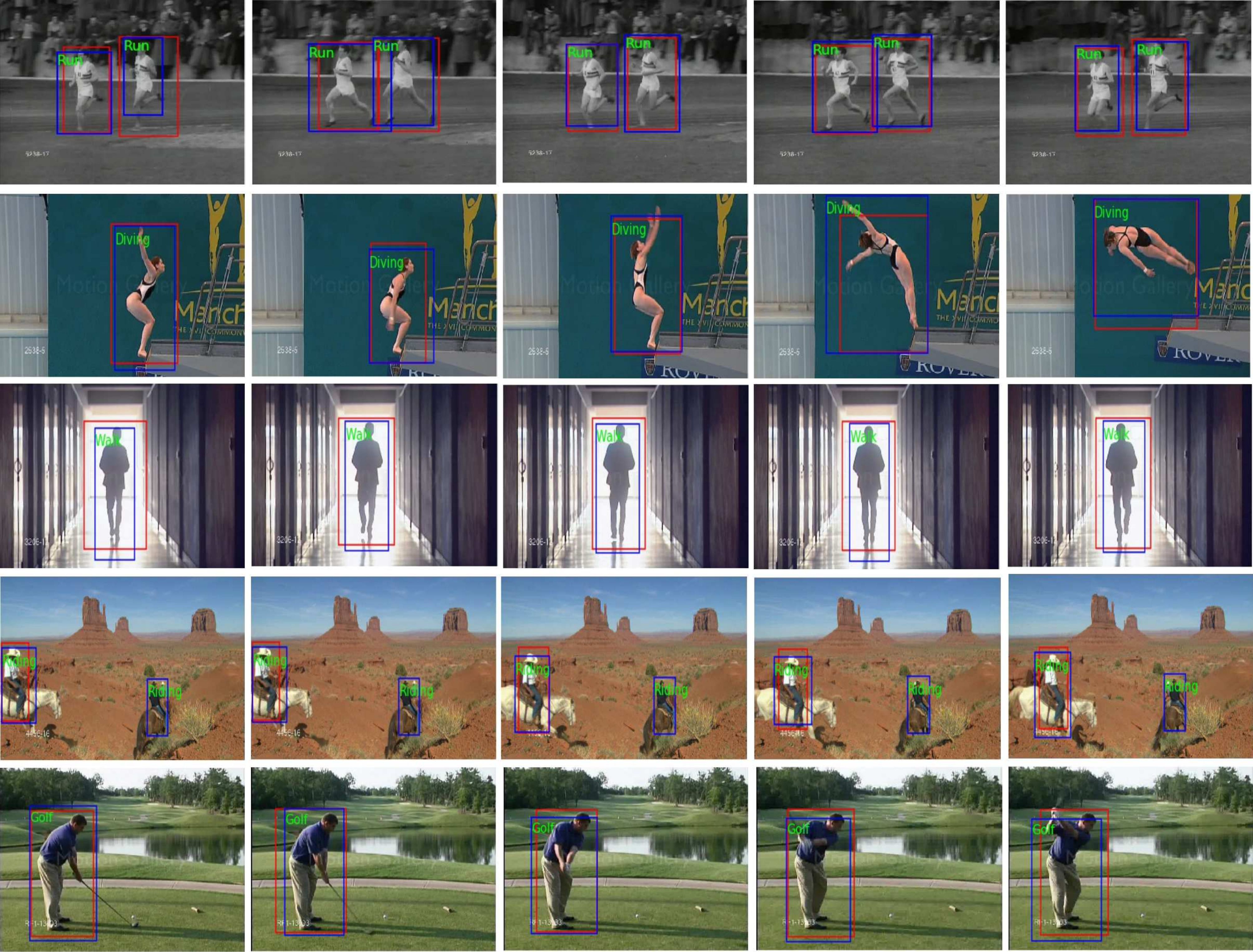}
\end{center}
   \caption{Visualization of Five detection examples from UCF-Sports dataset. Blue boxes indicate model detections and red boxes denote ground truths. The predicted label is on the top-left corner of each detection box.}
\label{fig:ucfvisual}
\end{figure*}

\begin{table*}
\caption{\textit{Frame-AP} of each class on UCF-Sports dataset with a IoU threshold $\sigma=0.5$.}
\centering
\begin{tabular}{p{3.3cm}p{1.0cm}p{0.6cm}p{1.0cm}p{0.9cm}p{0.9cm}p{0.5cm}p{1.0cm}p{1.0cm}p{0.8cm}p{0.7cm}p{0.6cm}}
\hline
  & Diving & Golf & Kicking & Lifting & Riding & Run & SkateB. & Swing1 & Swing2 & Walk & \textit{frame-mAP} \\
\hline
 Gkioxari \etal \cite{Gkioxari_2015_CVPR} & 75.8 & 69.3 & 54.6 & 99.1 & 89.6 & 54.9 & 29.8 & 88.7 & 74.5 & 44.7 & 68.1 \\
 Weinzaepfel \etal \cite{Weinzaepfel_2015_ICCV} & 60.71 & 77.55 & 65.26 & 100.00 & \textbf{99.53} & 52.60 & 47.14 & \textbf{88.88} & 62.86 & 64.44 & 71.9 \\
 Peng \etal \cite{peng2016multi} & 96.12 & 80.47 & 73.78 & 99.17 & 97.56 & 82.37 & 57.43 & 83.64 & 98.54 & 75.99 & 84.51 \\
 Hou \etal \cite{Hou_2017_ICCV} & 84.38 & 90.79 & 86.48 & 99.77 & 100.00 & 83.65 & 68.72 & 65.75 & \textbf{99.62} & \textbf{87.79} & 86.7 \\
 \hline
 Our DTN & \textbf{100.00} & \textbf{95.20} & \textbf{88.62} & \textbf{100.00} & 98.53 & \textbf{90.03} & \textbf{96.98} & 77.44 & 99.36 & 84.61 & \textbf{93.08} \\
\hline
\end{tabular}
\centering
\label{tab:2}
\end{table*}

\subsection{Comparison with Other Methods}

\subsubsection{Performance Comparison on UCF-Sports Dataset}Table~\ref{tab:2} shows our results of each class on UCF-Sports dataset. Our approach significantly outperforms other methods in the overall performance and most categories, e.g.\ \textit{Diving}, \textit{Kicking}, \textit{Lifting} and \textit{SkateBoarding}. Especially, our actor-centric tubelet feature helps action recognition for hard action classes, e.g, \textit{SkateBoarding}. We also make a comparison with recent methods and results are reported in Table~\ref{tab:3}. Overall, we outperform the state-of-the-art method \cite{Kalogeiton_2017_ICCV,RTPReccv2018} with a absolute margin of $5.38\%$ for \textit{frame-mAP} and $0.2\%$ for \textit{video-mAP} respectively. The results verify the effectiveness of our deformable actor-centric tube proposals, which benefit localizing actions with significant spatial motion.

In order to further validate the advantage of our method, we visualize five detection examples from UCF-Sports dataset in Figure~\ref{fig:ucfvisual}. Our method works well in complex cases such as those with multiple persons (first row) and large pose variations (second row). The false detections on the fourth row can be explained by high visual similarity with category \textit{Riding}. Moreover, our tube detections have high IoU overlaps with groundtruths, which is mainly due to actor-centric tube proposals with precise spatial-temporal context.


\begin{figure*}
\begin{center}
\includegraphics[width=1.0\linewidth]{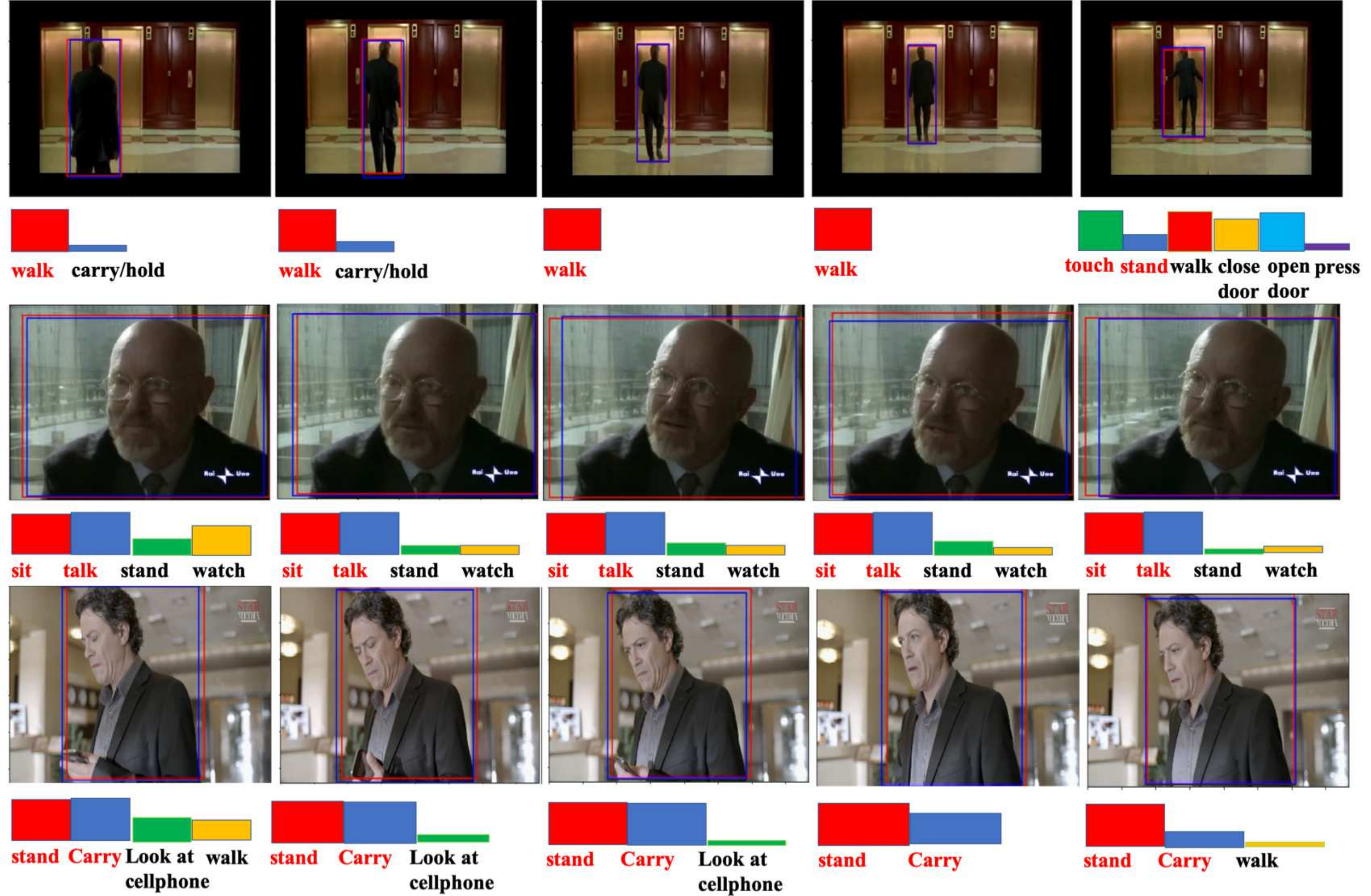}
\end{center}
   \caption{Visualization examples on AVA dataset. Blue boxes indicate model predictions and red boxes denote ground truths. The ground-truth labels are in red font.}
\label{fig:avavisual}
\end{figure*}

\begin{table}
\caption{Comparisons with the state of the arts on AVA dataset with \textit{frame-mAP}. The Iou threshold $\sigma$ is fixed to 0.5.}
\begin{center}
\begin{tabular}{lcccc}

\hline
  & Pretrained & \textit{frame-mAP} \\
 & dataset& $(\sigma=0.5)$ \\
\hline
Gu \etal \cite{Gu_2018_CVPR} (2D) & ImageNet & 14.2 \\
Li \etal \cite{RTPReccv2018} & ImageNet & 18.2 \\
Our DTN (2D) & ImageNet & \textbf{19.7} \\
\hline
Gu \etal \cite{Gu_2018_CVPR}(3D) & Kinetics-400 & 15.8  \\
Sun \etal \cite{actor-centric} & Kinetics-400 & 17.4  \\
Girdhar \etal \cite{action_transform} &   Kinetics-400 & 25.0\\
Wu \etal \cite{lfb2019} & Kinetics-400 & 26.8 \\
Feichtenhofer \etal \cite{slowfast} & Kinetics-600 & 27.3 \\
Our DTN (3D) & Kinetics-400 & 25.8 \\
Our DTN (3D) + LFB & Kinetics-400 & \textbf{27.7} \\
\hline
\end{tabular}
\end{center}
\label{tab:4}
\end{table}

\subsubsection{Performance Comparison on AVA Dataset}
AVA is a newly proposed dataset and there are only a handful of studies on it. We compare our results with the state-of-the-art methods \cite{Gu_2018_CVPR,actor-centric,RTPReccv2018,slowfast,lfb2019,action_transform} in Table~\ref{tab:4}. 
The rows of Table~\ref{tab:4} are split into two parts: methods with 2D backbone and 3D backbone. 
Among methods with 2D backbone, Gu \etal \cite{Gu_2018_CVPR} duplicated RPN detections of the key frame for all adjacent frames to form action tubes with limited spatial extent, while we can generate actor-centric tube proposals with our novel DTPN.  
The results indicate that we outperforms \cite{Gu_2018_CVPR,actor-centric} by a large margin. To have a fair comparison, we compare with the version in ~\cite{RTPReccv2018} which utilizes the same features with us, we can also observe that our method (2D) outperforms ~\cite{RTPReccv2018} by $1.5\%$ in terms of \textit{frame-mAP}.

\begin{figure*}
\begin{center}
\includegraphics[width=1.0\linewidth]{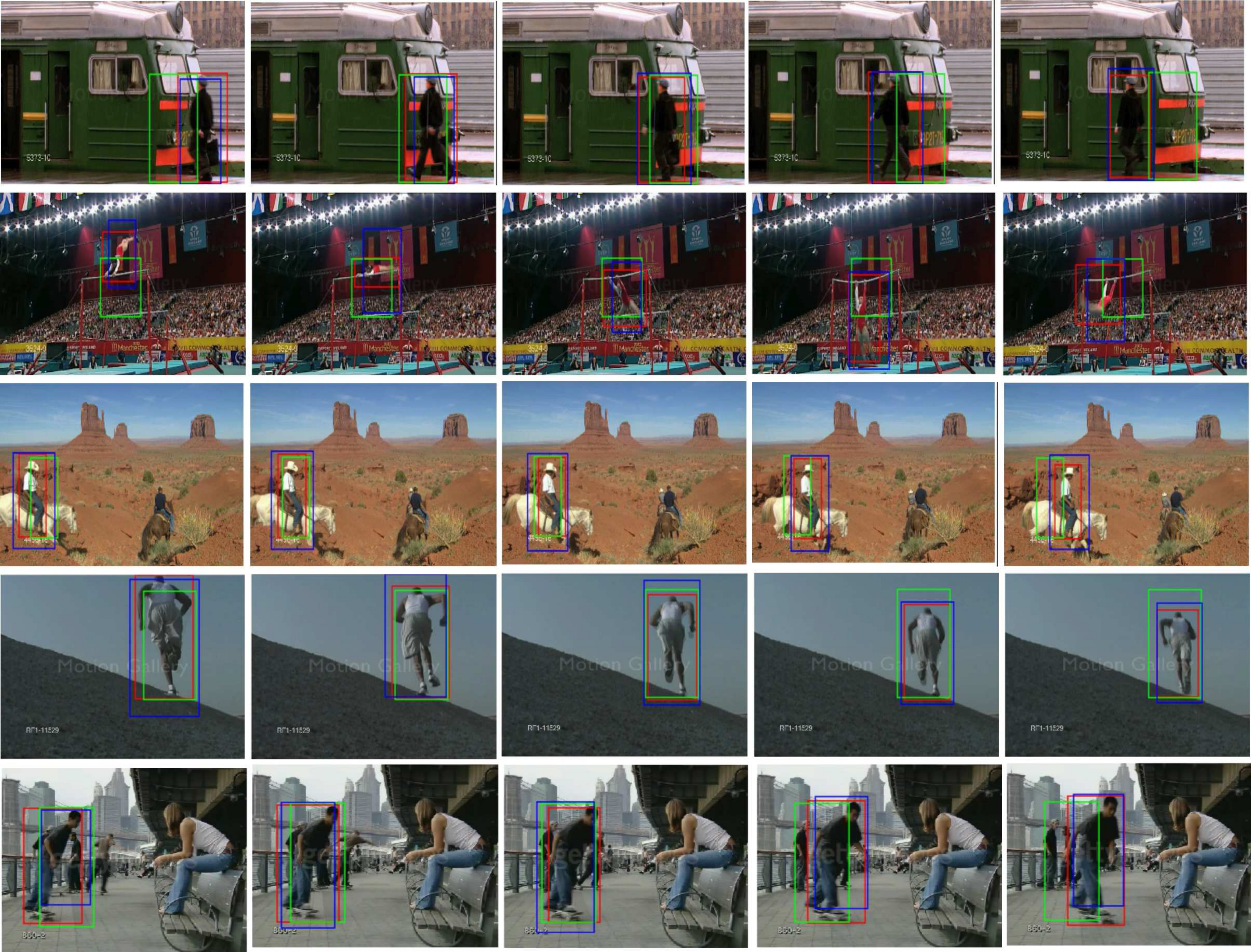}
\end{center}
   \caption{Visualization of linked tube examples with our DTPN. The green boxes represent region proposals of linked tubes and the red boxes means ground truths. The blue boxes indicates the fake anchor boxes.}
\label{fig:tubeexample}
\end{figure*}
For methods with 3D backbone, Feichtenhofer \etal~\cite{slowfast} propose a two-pathway network which treats spatial structures and temporal events separately. However, they only can integrate contents of short clips of 2-5 seconds. Wu \etal~\cite{lfb2019} resolve this with a long-term feature bank with supportive information extracted over the entire span of a video. They introduce a feature bank operator to compute interactions between long-term and short-term features. However, such long-term information may not relevant to the annotation of current key frame. Instead, Our DTN (3D) can integrate middle-term representation with our deformable linked tube which was verified to be complementary to LFB. Since the test augmentation strategies for existing methods are not always reported, we only compare with the results without test augmentation. It is worth mentioning that \cite{slowfast} pretrained their model on larger Kinetics-600 dataset which further boosts their performance. We achieve the state-of-the-art result on AVA dataset when combined with LFB.


Figure~\ref{fig:avavisual} shows three detection examples from AVA dataset. For example in the first row, our approach precisely localizes actor with its correct label \textit{walk}. Also, our approach can handle multi-label cases well such as \textit{sit} and \textit{talk} in the second row. To conclude, 
our system obtains the state-of-the-art result on AVA and we believe that fusing more scene features can further improve out detection performance. 

\section{Discussion}
\label{sec:discussion}
In this section, we make a further deeper dive into our system to validate the rationality of our fast tube linking algorithm.
\subsection{Visualization of Linked Tubes \& Anchor Cuboids}
\label{sec:2}
Kalogeiton \etal \cite{Kalogeiton_2017_ICCV} proposed to integrate temporal context by anchor cuboids with limited spatial extent. Our DTPN can generate actor-centric action tubes which contribute to precise action recognition and location regression. 
Table~\ref{tab:13} shows that our DTPN with linked tubes can outperforms its counterpart in most categories, especially with actors of fast motions. In order to further validate the effectiveness of our DTPN for actions even with large spatial motions, we select hard cases from UCF-Sports dataset for visualization.

Figure~\ref{fig:tubeexample} shows five examples of tube proposals and corresponding ground truth tubes in frame level. A fake anchor box, which is defined as average position of ground truth tube, is visualized in each frame to show the drawback of anchor cuboids with fixed spatial-extent. The top row shows a \textit{running} man with large spatial movement. Since our deformable tube proposal network adaptively links region proposals, the resulting tubes have high IoUs with groundtruths in each frame. However, the fake anchor box has a rather low IoU overlap with ground truths especially at the start and in the end. The same problem also exists in the second row with a fast-moving actor. Through the visualization, we verify the effectiveness of DTPN for cases with large spatial motions compared with anchor cuboids.


\subsection{Experiments on Top-K Selection}
\label{sec:ts}

Based on the observations that RPN generates rather reliable proposals and each region proposal of the best linked tube normally has a high objectness score, we modify Viterbi algorithm with top-K selection to accelerate the tube linking process. We argue that the linked tubes with or without top-K selection are intuitively similar. To quantify the effect of top-K selection, we propose a coselection rate $\gamma$ of tube proposal sets generated by methods with or without top-K selection
\begin{equation}
    \gamma=\frac{TP}{n}
\end{equation}
We select top n tube proposals generated with top-K selection for comparison. A tube proposal is assigned a positive label if it has a IoU higher than threshold $\theta$ with any tubes generated by the method without top-K selection. TP is the number of positive tubes among n tube proposals. $\gamma$ is averaged over the whole dataset. The criterion measures the distribution similarity between two tube proposal sets.
Table~\ref{tab:5} shows the comparison results of $\gamma$ with different $k$ and threshold $\theta$.

\begin{table}
\caption{Comparisons of coselection rate $\gamma$ on UCF-Sports dataset.}
\begin{center}
\begin{tabular}{lcccc}
\hline
 & $n=50$ & $n=100$ &$n=150$ &$n=200$\\
\hline
$\theta=0.7$ & 0.998 & 0.995 & 0.986 & 0.948\\
$\theta=0.8$ & 0.996 & 0.989 & 0.973 & 0.904\\
$\theta=0.9$ & 0.992 & 0.977 & 0.942 & 0.831\\
$\theta=1.0$ & 0.987 & 0.960 & 0.901 & 0.755\\
\hline
\end{tabular}
\end{center}
\label{tab:5}
\end{table}

We can see the tubes set with top-K selection has a high overlap with the tubes set without top-K selection. Specially, when $n=50$, even with a strong limitation $\theta=1.0$, we can still get a rather high coselection rate $\gamma=0.987$. According to our statistics, our improved linking algorithm can generate 
141 tubes averagely. The remaining tubes are linked only based on their objectness scores. 
The significant drop of $\gamma$ for $n=200$ compared with $n=150$ is due to lack of IoU restraints for those complementary tubes. The experiments validate the effectiveness of our fast linking algorithm with similar tube proposals and much lower time complexity.

\section{Conclusion}
\label{conclusion}
We propose Deformable Tube Network (DTN), a two-stage action localization architecture with a Deformable Tube Proposal Network (DTPN) and a Deformable Tube Recognition Network (DTRN). Compared with the methods based on anchor cuboids, DTPN generates deformable tube proposals by linking pre-frame region proposals with a fast tube linking algorithm. With actor-centric candidate action tubes, we use DTRN to perform action recognition and location regression with a 3D convolutional network with skip connections to integrate spatio-temporal context. Our experiments validate the effectiveness of our method. Moreover, we achieve the state-of-the-art results on both UCF-Sports and AVA datasets.

{\small
\bibliographystyle{ieee_fullname}
\bibliography{egbib}
}

\end{document}